# Advances in Artificial Intelligence:
# Deep Intentions, Shallow Achievements

Emanuel Diamant
VIDIA-mant/ POB 933, Kiriat Ono 5510801, Israel
emanl.245@gmail.com

**Abstract:** Over the past decade, AI has made a remarkable progress due to recently revived Deep Learning technology. Deep Learning enables to process large amounts of data using simplified neuron networks that simulate the way in which the brain works. At the same time, there is another point of view that posits that brain is processing information, not data. This duality hampered AI progress for years. To provide a remedy for this situation, I propose a new definition of information that considers it as a coupling between two separate entities – physical information (that implies data processing) and semantic information (that provides physical information interpretation). In such a case, intelligence arises as a result of information processing. The paper points on the consequences of this turn for the AI design philosophy.

## I. INTRODUCTION

Artificial Intelligence was invented at the Dartmouth College meeting in 1956. Four brilliant scientists (J. McCarthy, M.L. Minsky, N. Rochester, and C.E. Shannon) have worked out and put into existence the AI idea. However, despite the famous fathers, AI research has never been skilful enough even to formulate its main goals. The fathers have failed to assess the complexity of the task and mutual contradictions between its basic constituents. (Recall the story how Minsky proposed to hire a student to solve the problem of vision during the student's summer vocations).
The core problem was the lack of clear and well-defined research fundamentals: First, what is intelligence and how it should be defined? The multiplicity of available suggestions (70+ are collected in [1]) is only a sign of how tough and complicated the problem is.
In absence of a proper definition, the AI research community has adopted concepts widespread in the mid-1950s: "brain as a computer" metaphor was the most popular at that time; computer has been accepted as a data processing device, so intelligence has been assumed as a data processing implement. "Computational" have become almost all other scientific fields: Computational biology, Computational genomics, Computational neuroscience, Computational linguistics, and so on. In such a case, why not Computational intelligence?
Because of this fashion, all breakthrough accomplishments in AI today [2] are related to advances in data processing techniques and algorithms: Deep learning is the most prominent among them. Wikipedia defines Deep learning as a set of machine learning algorithms that attempt to model high-level abstractions in data by using model architectures composed of multiple non-linear transformations, [3]. Deep learning software works by filtering data through a hierarchical, multilayered network of simulated neurons that are individually simple but can exhibit complex behavior when linked together, [3].
However, contemporary brain science theories insist that nervous system, which is the core of human intelligence, is busy with processing information, not data. What is the difference? A clear-cut answer cannot be proposed because no one really knows what information is. A consensus definition of it does not exist, therefore the only option is to set up the mind and to try to get the desired definition in a new way.

## II. SO, WHAT IS INFORMATION?

As it was said, a proper definition of information does not exist. Therefore, I would like to propose my own one. It is an extended version of Kolmogorov's mid-60s definition [4], which now can be expressed in a following way:
**"Information is a linguistic description of structures observable in a given data set".**
A digital image would be a suitable testbed for our definition analysis. An image is a two-dimensional set of data elements called pixels. In an image, pixels are distributed not randomly, but due to the similarity in their physical properties, they are naturally grouped into some clusters or clumps. I propose to call these clusters **primary or physical data structures**.
In the eyes of an external observer, the primary data structures are further arranged into more larger and complex assemblies (usually called "visual objects"), which I propose to call **secondary data structures**. These secondary structures reflect human observer's view on the primary data structures arrangements, and therefore they could be called **meaningful or semantic data structures**. While formation of primary data structures is guided by objective (natural, physical) properties of the data, ensuing formation of secondary structures is a subjective process guided by human habits and customs.
As it was said, **Description of structures observable in a data set should be called "Information".** In this regard, two types of information must be distinguished – **Physical Information and Semantic Information**. They are both language-based descriptions; however, physical information can be described with a variety of languages (recall that mathematics is also a language), while semantic information can be described only by means of the natural human language. (More details on the subject can be find in [5]).
Every information description is a top-down evolving coarse-to-fine hierarchy of descriptions representing various levels of description complexity (various levels of description details). Physical information hierarchy is located at the lowest level of the semantic hierarchy. The process of sensor data interpretation is reified as a process of physical information extraction from the input data, followed by an attempt to associate this physical information with physical information already retained at the lowest level of a semantic hierarchy. If such association is reached, the input physical information becomes related (via the physical information retained in the system) with a relevant linguistic term, with a word that places the physical information in a context of a phrase, which provides the semantic interpretation of it. In such a way, the input physical information becomes named with an appropriate linguistic label and framed into a suitable linguistic phrase (and further – in a story, a tale, a narrative), which provides the desired meaning for the input physical information.

## III. RETHINKING INTELLIGENCE

In the light of the above elucidation, the definition of intelligence, for the first time, can be put forward as such: **Intelligence is an ability to process information (in particular semantic information)**. Adoption of this definition naturally raise a question: Is the human brain the only proper means to facilitate this purpose? The answer is **obviously not**! Bacteria and amoebas exhibit intelligent behavior (intentional external world interaction, which certainly requires information processing) without any sign of a brain or a nervous system. Thus, it must be emphasized that the new definition is equally applicable to human beings, to all other living beings, and to all natural and artificial systems as well.

That poses a challenge – AI design was always aimed to mimic human brain intelligence. Now it has to rely on semantic information processing only. The human brain is not needed more for such an activity. That is, we are on the verge of brainless intelligence modeling, which will certainly affect the future AI design.

## IV. CONCLUSION

Contemporary science witnesses a paradigm shift from data-based computational approach to information-based cognitive approach.

The computational paradigm that emerged in the second half of the past century has been generally accepted as the prevalent paradigm of that time scientific community. For that reason, we have "computational intelligence" as well as "computational biology" or "computational linguistics". The brain has become regarded as a computing device, that is, a device aimed at number crunching and data manipulation. Therefore, even today, all AI algorithms are devised to process and to operate data. All breakthrough achievements of the last time are data-processing implements, (Deep Learning is assumed as the most prominent among them).

At the same time, it is generally accepted that the brain is an information-processing engine. The contradiction between the two definitions can be explained by the peculiarities of scientific development in the past century. It is generally accepted that Intelligence of a living being is expressed in his behavior, that is, in his interaction and communication with the environment. Claude Shannon (one of the AI founders) in his seminal "Mathematical Theory of Communication" [6], has defined that what is being conveyed in a communication process is Information. Shannon was aware that this definition of information is applicable only to data communication, and has nothing to say about the meaning of the conveyed message. Shannon's Information Theory has become widespread and popular in almost all fields of scientific research, despite its shortfalls in message semantics handling. The results of this deficiency are well recognized (over the all AI history) – they have derailed AI research permanently and forever.

This paper is aimed to help the AI researchers to understand the roots of their permanent failures. I have introduced here a new definition of information that will certainly help to avoid in the future the prior mistakes.